\documentclass[journal]{IEEEtran}
\usepackage{amsmath}
\usepackage{algorithm}
\usepackage{algorithmic}
\usepackage{graphicx}
\usepackage{multirow}
\usepackage{tabularx}
 
\usepackage{xcolor}
\usepackage{array}
\usepackage{amssymb}
\usepackage{bbding}
\usepackage{subfigure}
\usepackage{diagbox}
\usepackage{rotating}
\usepackage{bm}
\usepackage{tikz} 
\usepackage{booktabs}
\usepackage{cite}
\usepackage{CJK}
\usepackage{overpic}
\usepackage[colorlinks,
            linkcolor=red,
            anchorcolor=blue,
            citecolor=green
            ]{hyperref}

\newcommand{\addFig}[1]{}
\newcommand{\addFigs}[1]{}

\usepackage{subfigure}
\newcommand{\etal}{\textit{et~al}.~}
\newcommand{\ie}{\textit{i}.\textit{e}.,~}
\newcommand{\eg}{\textit{e}.\textit{g}.,~}

\usepackage{balance}
\usepackage{cleveref}
\crefformat{section}{\S#2#1#3} 
\crefformat{subsection}{\S#2#1#3}
\crefformat{subsubsection}{\S#2#1#3}

\ifCLASSINFOpdf
\else
\fi

\begin{document}
%
\title{Lightweight Salient Object Detection in \\Optical Remote-Sensing Images via \\Semantic Matching and Edge Alignment}
%
%
%

\author{Gongyang~Li,
	Zhi~Liu,~\IEEEmembership{Senior Member,~IEEE},
	Xinpeng~Zhang,~\IEEEmembership{Member,~IEEE},
	and~Weisi~Lin,~\IEEEmembership{Fellow,~IEEE}

\thanks{Gongyang Li, Zhi Liu, and Xinpeng Zhang are with Key Laboratory of Specialty Fiber Optics and Optical Access Networks, Joint International Research Laboratory of Specialty Fiber Optics and Advanced Communication, Shanghai Institute for Advanced Communication and Data Science, Shanghai University, Shanghai 200444, China, and School of Communication and Information Engineering, Shanghai University, Shanghai 200444, China (email: ligongyang@shu.edu.cn; liuzhisjtu@163.com; xzhang@shu.edu.cn).}
\thanks{Weisi Lin is with the School of Computer Science and Engineering, Nanyang Technological University, Singapore 639798 (e-mail: wslin@ntu.edu.sg).}
\thanks{\textit{Corresponding authors: Zhi Liu and Xinpeng Zhang.}}
}

\markboth{IEEE TRANSACTIONS ON GEOSCIENCE AND REMOTE SENSING}%
{Shell \MakeLowercase{\textit{et al.}}: Bare Demo of IEEEtran.cls for IEEE Journals}

\maketitle

\begin{abstract}
Recently, relying on convolutional neural networks (CNNs), many methods for salient object detection in optical remote sensing images (ORSI-SOD) are proposed.
However, most methods ignore the huge parameters and computational cost brought by CNNs, and only a few pay attention to the portability and mobility.
To facilitate practical applications, in this paper, we propose a novel lightweight network for ORSI-SOD based on \underline{s}emantic matching and \underline{e}dge \underline{a}lignment, termed \emph{SeaNet}.
Specifically, SeaNet includes a lightweight MobileNet-V2 for feature extraction, a dynamic semantic matching module (DSMM) for high-level features, an edge self-alignment module (ESAM) for low-level features, and a portable decoder for inference.
First, the high-level features are compressed into semantic kernels.
Then, semantic kernels are used to activate salient object locations in two groups of high-level features through dynamic convolution operations in DSMM.
Meanwhile, in ESAM, cross-scale edge information extracted from two groups of low-level features is self-aligned through $L_2$ loss and used for detail enhancement.
Finally, starting from the highest-level features, the decoder infers salient objects based on the accurate locations and fine details contained in the outputs of the two modules.
Extensive experiments on two public datasets demonstrate that our lightweight SeaNet not only outperforms most state-of-the-art lightweight methods but also yields comparable accuracy with state-of-the-art conventional methods, while having only 2.76M parameters and running with 1.7G FLOPs for 288$\times$288 inputs.
Our code and results are available at https://github.com/MathLee/SeaNet.
\end{abstract}

\begin{IEEEkeywords}
Optical remote sensing image, lightweight salient object detection, semantic matching, edge alignment.
\end{IEEEkeywords}

\IEEEpeerreviewmaketitle

\section{Introduction}
\IEEEPARstart{S}{alient} object detection (SOD) aims at imitating the human vision system to quickly locate the objects/areas that attract the most attention~\cite{19CRMCO}.
As an important preprocessing step, the success of SOD has promoted the development of many fields, such as image quality assessment~\cite{16SODIQA,19SGDNet}, object segmentation~\cite{LGY2019,LGY2021PFOS}, and object tracking~\cite{20TRACK}.
Different from most SOD methods proposed for single image~\cite{2019sodsurvey}, RGB-D/T image~\cite{20ICNet,20CMWNet}, and video~\cite{WWG19Video} photographed in natural scenes, in this paper, we focus on SOD in optical remote sensing images, or ORSI-SOD for short. 
Following the technical trend of SOD in natural scene images (NSI-SOD)~\cite{2019sodsurvey}, we are committed to addressing ORSI-SOD based on convolutional neural networks (CNNs)~\cite{1989CNN}.

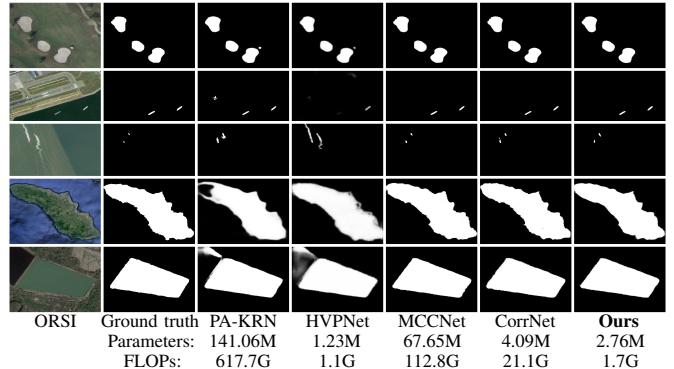
\begin{figure}[t!]
  \centering
  \scriptsize
  \begin{overpic}[width=1\columnwidth]{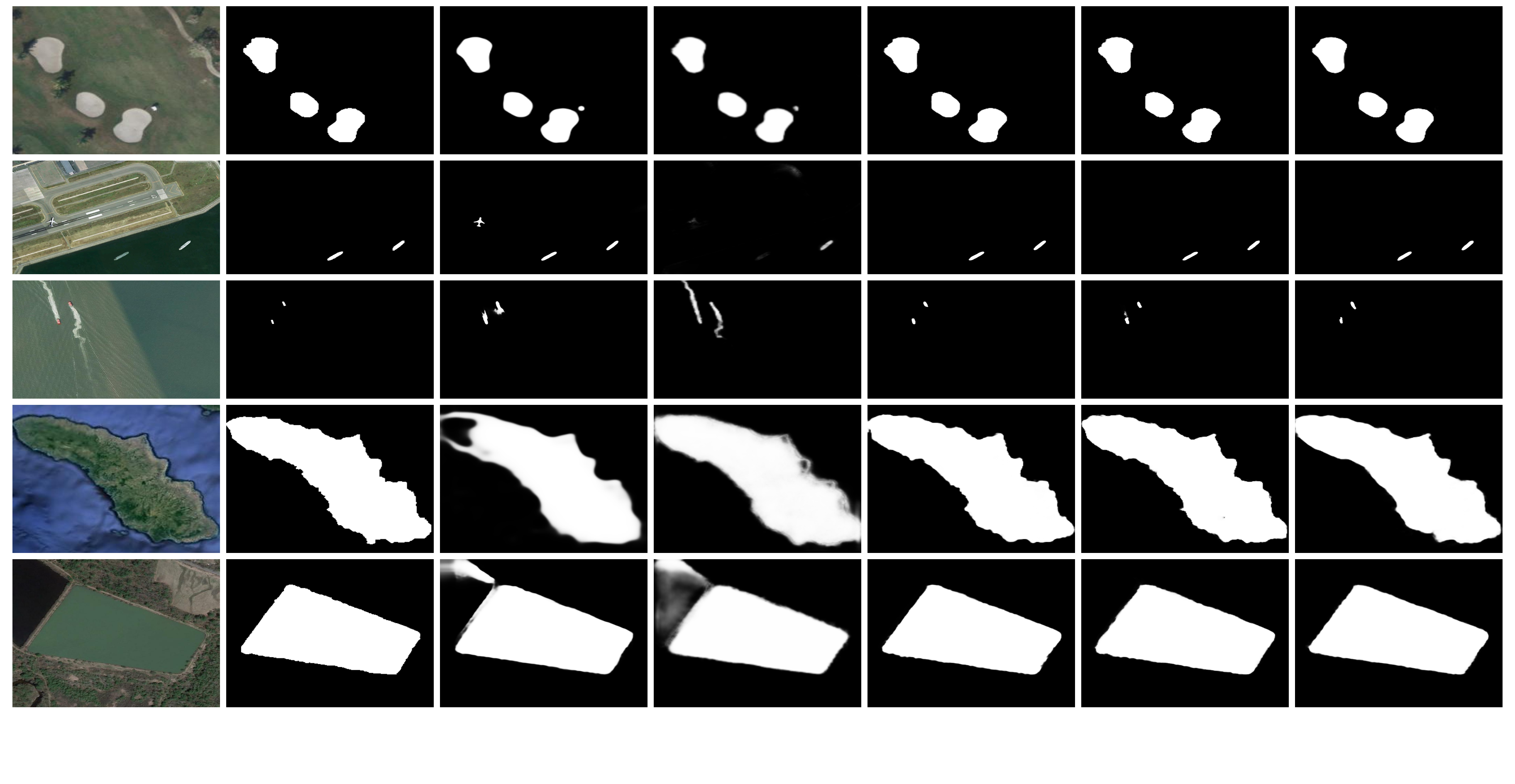}
      \put(3.5,2.8){ ORSI}
      \put(13.5,2.8){ Ground truth }   
      \put(14.5,-0.2){ Parameters:}
      \put(16.9,-3.2){ FLOPs:}
            
      \put(29.8,2.8){ PA-KRN}
      \put(29.7,-0.2){ 141.06M}
      \put(30.7,-3.2){ 617.7G}
      
      \put(44.2,2.8){ HVPNet}
      \put(45.17,-0.2){ 1.23M}
      \put(46.17,-3.2){ 1.1G}
      
      \put(58.05,2.8){ MCCNet}
      \put(58.85,-0.2){ 67.65M}
      \put(59,-3.2){ 112.8G}
    
      \put(72.6,2.8){ CorrNet}
      \put(73.6,-0.2){ 4.09M}   
      \put(73.8,-3.2){ 21.1G}
      
      \put(88.2,2.8){ \textbf{Ours}}
      \put(87.8,-0.2){ 2.76M}
       \put(88.7,-3.2){ 1.7G}
  \end{overpic}
  \caption{Saliency maps produced by four types of methods and our method on ORSIs. PA-KRN~\cite{2021PAKRN} is an NSI-SOD method, HVPNet~\cite{2021HVPNet} is a lightweight NSI-SOD method, MCCNet~\cite{2022MCCNet} is an ORSI-SOD method, and CorrNet~\cite{2022CorrNet} is a lightweight ORSI-SOD method.
  Please zoom-in for details.
    }\label{fig:example}
\end{figure}

In the era of deep learning, numerous CNN-based NSI-SOD methods have been proposed, and the detection accuracy has been significantly improved.
Among these methods, the classic encoder-decoder structure~\cite{2015Unet} is the most general and effective structure, and is often accompanied with ingenious strategies such as deep supervision~\cite{2017DSS}, gate mechanism~\cite{2020GateNet}, edge assistance~\cite{2019EGNet,2020ITSD}, progressive architecture~\cite{2021PAKRN}, \textit{etc}.
Although NSI-SOD methods cannot directly overcome the issue of complex scenes of ORSIs (as PA-KRN~\cite{2021PAKRN} shown in the third column of Fig.~\ref{fig:example}), the strategies contained therein lay the foundation for CNN-based ORSI-SOD methods. 
The specialized methods for ORSI-SOD take into account the properties of salient objects and scenes in ORSIs.
For example, LVNet~\cite{2019LVNet} and EMFINet~\cite{2022EMFINet} take ORSIs with multiple resolutions as inputs to overcome the problem of variable sizes of salient objects.
MCCNet~\cite{2022MCCNet} comprehensively integrates foreground, background, edge, and global information to deal with the complex background of ORSIs, producing good saliency maps as shown in the fifth column of  Fig.~\ref{fig:example}.

However, the above methods may fall into the dilemma of huge amount of parameters and computational cost, such as the parameters and FLOPs of PA-KRN and MCCNet listed in Fig.~\ref{fig:example}.
To address this issue, the lightweight SOD methods are gradually emerging.
Compared with PA-KRN, the pioneer of lightweight NSI-SOD method HVPNet~\cite{2021HVPNet} reduces the amount of parameters and computational cost by hundreds of times.
But HVPNet is also stuck by ORSIs, as shown in the fourth column of Fig.~\ref{fig:example}.
For the lightweight ORSI-SOD method, as shown in the penultimate column of Fig.~\ref{fig:example}, CorrNet~\cite{2022CorrNet} significantly reduces the amount of parameters and achieves good performance, but still consumes a lot of computational cost.

Driven by the aforementioned observation, in this paper, we propose a novel \underline{s}emantic matching and \underline{e}dge \underline{a}lignment based ORSI-SOD method, termed \emph{SeaNet}, which aims to be more lightweight than CorrNet, while generating competitive performance.
As we all know, the features extracted by the feature extraction network can be divided into low-level and high-level features, where the former contains detail and texture information and the latter contains semantic and location information.
Accordingly, the main idea of SeaNet is to explore high-level and low-level features with different strategies in the encoder-decoder structure.

Specifically, we propose a \emph{Dynamic Semantic Matching Module} to implement the semantic matching in high-level features, that is, first compress semantic information and then match them with the high-level features to perceive the location of salient objects.
We also propose an \emph{Edge Self-Alignment Module} for edge alignment in low-level features, that is, align cross-scale edge information extracted from low-level features to correct edge errors and use them to enhance features.
For efficiency, we adopt MobileNet-V2~\cite{MobileNet2} as the backbone and the depthwise separable convolution~\cite{MobileNet1,MobileNet2} as the basic convolution component to control the amount of parameters and computational cost.
In this way, our SeaNet has only 2.76M parameters, runs with 1.7G FLOPs, and can generate accurate saliency maps, as shown in the rightmost column of Fig.~\ref{fig:example}.
Concretely, compared with state-of-the-art lightweight methods, our SeaNet is competitive in detection accuracy, while compared with state-of-the-art conventional methods, our SeaNet is competitive in computational complexity.

Our main contributions are threefold:
\begin{itemize}
\item We explore high-level and low-level features of MobileNet-V2 with different strategies, and propose a novel lightweight network for ORSI-SOD based on semantic matching and edge alignment, termed \emph{SeaNet}, which has only 2.76M parameters and runs with 1.7G FLOPs for a 288$\times$288 image.

\item We propose a \emph{Dynamic Semantic Matching Module} for high-level semantic features.
DSMM perceives the location of salient objects through dynamic convolutions with semantic kernels, which not only improves the flexibility of feature interaction but also effectively reduces the amount of parameters.
Moreover, it performs channel-wise correlation to activate the channel-wise interaction.

\item We propose an \emph{Edge Self-Alignment Module} for low-level detail features.
ESAM focuses on extracting edge information for detail enhancement, and aligns cross-scale edge information through $L_2$ loss to correct edge errors.
Like DSMM, it also introduces channel-wise correlation.

\end{itemize}

We arrange the remainder of this paper as follows.
In Sec.~\ref{sec:related}, we review the CNN-based SOD methods for NSIs and ORSIs.
In Sec.~\ref{sec:OurMethod}, we elaborate our SeaNet.
In Sec.~\ref{sec:exp}, we present comprehensive experimental results.
In Sec.~\ref{sec:con}, we give the conclusion.

\section{Related Work}
\label{sec:related}

\subsection{CNN-based Salient Object Detection in NSIs}
\label{sec:NSI_SOD}
Salient object detection in natural scene images~\cite{2019sodsurvey} has achieve remarkable success, especially CNN-based methods.
In~\cite{2017DSS}, Hou~\etal creatively introduced the pioneer deep supervision into NSI-SOD, which significantly enhances the representation of multi-scale features for salient objects.
This method effectively improves the detection accuracy and has a profound impact on subsequent CNN-based methods.
Hu~\etal\cite{2018RADF} and Deng~\etal\cite{2018R3Net} focused on the recurrent mechanism.
The former first concatenates multi-level features, and then combines them with features from different levels.
The latter alternatively uses the low-level features and high-level features.
Besides, some researchers were interested in edge/boundary information.
Feng~\etal\cite{2019AFNet} proposed a boundary-enhanced loss to improve the completeness of object boundaries, and combined it with deep supervision.
Qin~\etal\cite{2019BASNet} proposed a hybrid loss, which integrates BCE, SSIM and IoU losses, to segment salient object with fine structures and clear boundaries.
Zhao~\etal\cite{2019EGNet} modeled the explicit edge through edge supervision to preserve the salient object boundaries.
Zhou~\etal\cite{2020ITSD} proposed the saliency to contour and contour to saliency strategy for fast saliency detection.
Lee~\etal\cite{2022TRACER} proposed an attention guided tracing module to highlight salient objects with explicit edges.

In addition to the above classic strategies, Pang~\etal\cite{2020MINet} enhanced the interaction of multi-scale features for NSI-SOD.
Chen~\etal\cite{2020GCPA} considered the low-level, high-level and global information to improve the completeness of saliency map.
Xu~\etal\cite{2021PAKRN} utilized the global localization and local segmentation policy in the knowledge review network to avoid salient information dilution.
In~\cite{2021SUCA}, Li~\etal captured both the multi-receptive-field information of features and the complementary information of cross-level features.
Qiu~\etal\cite{2022RCSB} extended the atrous spatial pyramid pooling, and embedded the channel and spatial attention into it to explore information dependencies in space and channel.

Although the above CNN-based methods achieve excellent performance in NSI-SOD, they cannot effectively handle the unique properties of ORSIs and often generate unsatisfactory saliency maps.
Furthermore, they usually focus on accuracy and ignore computational complexity.
Nonetheless, their strategies inspire our approach, such the classic deep supervision, edge assistance, and differential feature processing.

\subsection{CNN-based Salient Object Detection in ORSIs}
\label{sec:ORSI_SOD}
Salient object detection in optical remote sensing images is a rising star in SOD community, and recently numerous CNN-based methods are proposed.
As a pioneer, Li~\etal\cite{2019LVNet} proposed the first CNN-based ORSI-SOD method, named LVNet, in which a two-stream pyramid module cooperates with an encoder-decoder module with nested connections to perceive salient objects of different sizes.
Moreover, Li~\etal\cite{2020PDFNet} explored the interactions of cross-level features for ORSI-SOD.
In~\cite{2021SARNet}, Huang~\etal first roughly located salient objects in the semantic guided decoder, and then refined the coarse saliency map in a top-down manner.
Cong~\etal\cite{2022RRNet} designed the relational reasoning encoder for high-level features, and inferred salient objects in a multiscale attention decoder.
Li~\etal\cite{2022ACCoNet} adopted two groups of adjacent features to assist the current features, capturing contextual information to overcome challenging ORSI scenarios.

Similar to NSI-SOD, some researchers introduced edge/boundary information into ORSI-SOD.
Zhang~\etal\cite{2021DAFNet} constructed a multi-task structure to predict edge map and saliency map simultaneously.
Tu~\etal\cite{2022MJRBM} and Zhou~\etal\cite{2022EMFINet}, following~\cite{2019EGNet}, extracted boundary information based on low-level and high-level features to preserve boundaries of salient objects in two decoders.
Li~\etal\cite{2022MCCNet} integrated edge with foreground, background, and global information, and took full account of the complementarity between these information to adapt to ORSIs.

The above specialized methods for ORSI-SOD achieve satisfactory performance.
However, they usually come with a large number of parameters and heavy computational cost, which are unfriendly to aerospace equipments and prevent practical applications.
To this end, we propose SeaNet based on MobileNet-V2~\cite{MobileNet2} and two lightweight but effective modules, which are friendly to mobile devices while achieving competitive performance.

\subsection{Lightweight Salient Object Detection}
\label{sec:Light_SOD}
Lightweight SOD is a newly emerging task, and is first explored in NSI-SOD.
Gao~\etal\cite{2020CSNet} proposed a flexible self-adaptive convolutional layer with strong multi-scale representation abilities and constructed an extremely lightweight network (\ie CSNet) for NSI-SOD.
Liu~\etal\cite{2021HVPNet} proposed a hierarchical visual perception (HVP) module based on dense connections, and built a lightweight HVPNet on HVP modules and residual attention to effectively learn multi-scale contexts.
Meanwhile, Liu~\etal\cite{2021SAMNet} proposed a stereoscopically attentive multi-scale (SAM) module, and built a lightweight SAMNet for multi-level and multi-scale learning.
To sum up, the above three works focus on learning effective multi-level and multi-scale information for SOD, and build strong feature extraction backbones to achieve lightweight NSI-SOD.
However, they do not further process and enhance the extracted features at different levels, and directly infer salient objects based on these features.
Moreover, they deal with NSIs rather than ORSIs, lacking pertinence.
Therefore, in this paper, we focus on making good use of features at different levels of existing feature extraction backbones and developing specific lightweight and effective modules for ORSI features, enabling more effective dedicated ORSI-SOD solution.

For ORSI-SOD, Li~\etal\cite{2022CorrNet} proposed the first lightweight method, \ie CorrNet.
They lightened the vanilla VGG-16~\cite{2014VGG16ICLR} for efficient feature extraction, and adopted the coarse-to-fine strategy to detect salient objects in ORSIs with dense lightweight refinement blocks.
The parameters of CorrNet were greatly reduced to only 4.09M, but its computational cost was still very large, with 21.1G FLOPs.
For RGB-D SOD, Wu~\etal\cite{2021MobileSal} proposed the first lightweight method, \ie MobileSal, which is based on MobileNet-V2~\cite{MobileNet2}.

Inspired by~\cite{2021MobileSal}, in SeaNet, we adopt MobileNet-V2 as the backbone to overcome the issue of large computational cost of the existing lightweight ORSI-SOD method, \ie CorrNet.
Moreover, to reduce the amount of parameters, we propose two lightweight and effective modules, \ie DSMM and ESAM.
DSMM extends dynamic convolution~\cite{DyConv2020} to dynamic depthwise convolution, and ESAM corrects cross-scale edge features in a self-alignment way.

\section{Proposed Method}
\label{sec:OurMethod}
In this section, we elaborate the proposed lightweight SeaNet.
In Sec.~\ref{sec:Overview}, we introduce the network overview of proposed SeaNet.
In Sec.~\ref{sec:DSMM} and Sec.~\ref{sec:ESAM}, we elaborate two lightweight modules, \ie DSMM and ESAM, respectively.
In Sec.~\ref{sec:Decoder_and_LossFunction}, we depict the decoder and loss function.


\begin{figure*}
	\centering
	\begin{overpic}[width=1\textwidth]{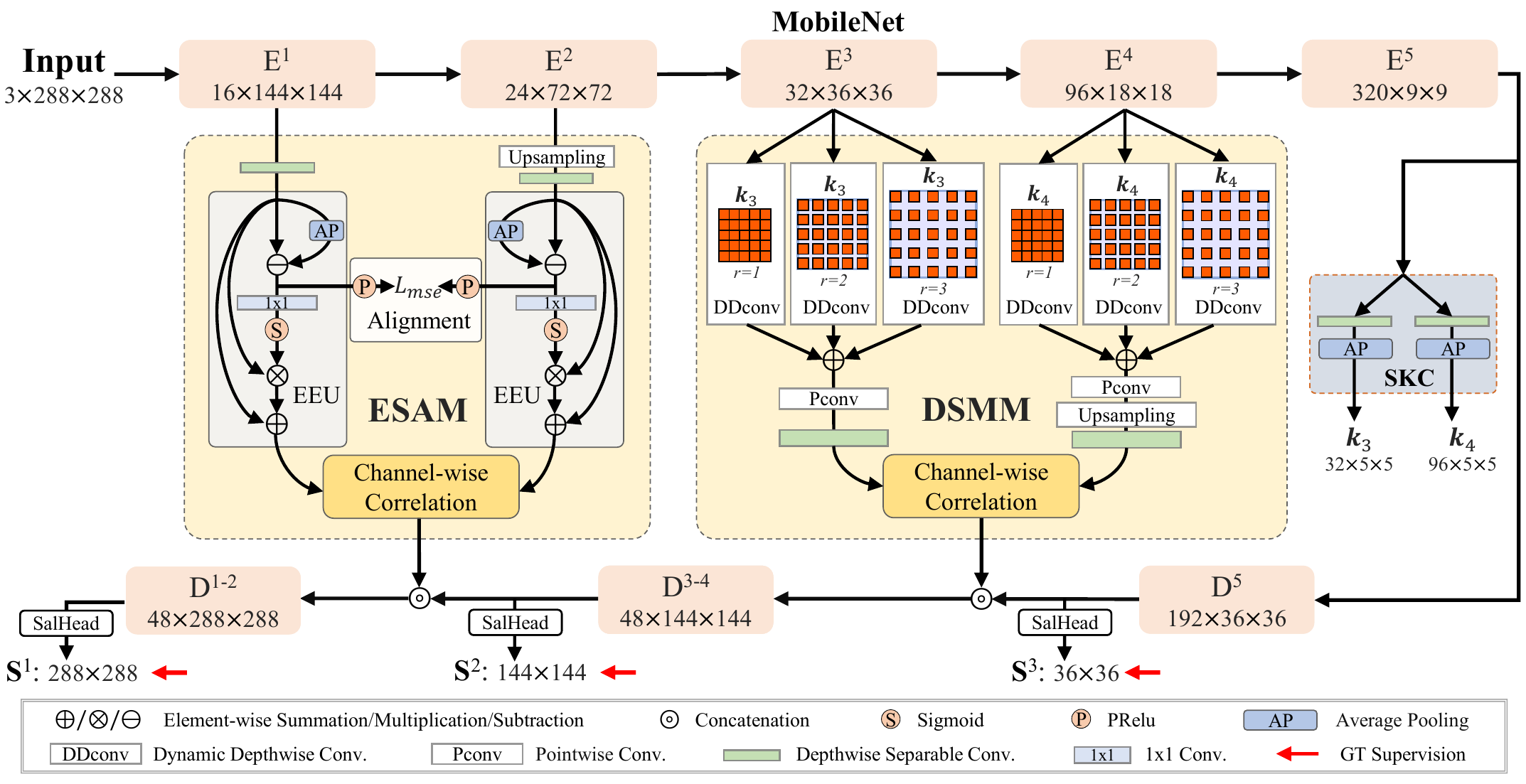}
    \end{overpic}
	\caption{Pipeline of the proposed lightweight SeaNet, which follows the encoder-decoder architecture.
	First, MobileNet-V2~\cite{MobileNet2} extracts basic feature embeddings from the input, resulting in five-level features.
	Next, we get two semantic kernels in the Semantic Knowledge Compression (SKC) unit.
	Semantic kernels are used to convolve with high-level features in the Dynamic Semantic Matching Module (DSMM) for location activation of salient objects.
	Then, we perform the channel-wise correlation~\cite{2016CoAtt,2022COSNet} in DSMM.
	Meanwhile, in the Edge Self-Alignment Module (ESAM), we adopt Edge-based Enhancement Units (EEUs) to extract edge features from low-level features for detail enhancement, and align edge features via ${L}_{mse}$ loss (\ie ${L}_{2}$ loss).
	Channel-wise correlation is also added to ESAM.
	Finally, we infer salient objects in the decoder based on the highest-level features and the outputs of DSMM and ESAM, and obtain the output saliency map $\mathbf{S}^{\textrm{1}}$.
    }
    \label{fig:Framework}
\end{figure*}

\subsection{Network Overview}
\label{sec:Overview}
As shown in Fig.~\ref{fig:Framework}, the proposed lightweight SeaNet is based on the encoder-decoder structure commonly used in SOD~\cite{20ICNet,21HAINet,20CMWNet,2021AGCNet}.
SeaNet includes an encoder, a Semantic Knowledge Compression (SKC) unit, a Dynamic Semantic Matching Module (DSMM), an Edge Self-Alignment Module (ESAM), and a lightweight decoder.
It first performs semantic matching for location activation of salient objects, and then performs edge alignment for detail enhancement.

The input size of our SeaNet is $3\!\times\!288\!\times\!288$.
For the encoder of our SeaNet, we adapt the lightweight MobileNet-V2~\cite{MobileNet2}, that is, we keep the first seventeen inverted residual bottlenecks and truncate the last three layers, \ie two convolution layers and one average pooling layer.
We divide MobileNet-V2 into five blocks based on the first, third, sixth, thirteenth and last bottlenecks, denoted as E$^{t}$ ($t=1,2,3,4,5$).
The output five-level features are denoted as $\boldsymbol{f}^{t}_{\rm e} \in \mathbb{R}^{c_t\!\times\!h_t\!\times\!w_t}$, where $h_t$ and $w_t$ are $\frac{288}{2^{t}}$, and $c_t \in \{16,24,32,96,320\}$.
We explore the generated high-level and low-level features with different strategies for ORSI-SOD.
For the high-level features, we first compress the highest-level features $\boldsymbol{f}^{5}_{\rm e}$ into two semantic kernels, \ie $\boldsymbol{k}_{3}$ and $\boldsymbol{k}_{4}$, in the Semantic Knowledge Compression (SKC) unit, and then use them as kernels for dynamic depthwise convolutions~\cite{DyConv2020,MobileNet1} to respectively convolve with two groups of high-level features, \ie $\boldsymbol{f}^{3}_{\rm e}$ and $\boldsymbol{f}^{4}_{\rm e}$, in DSMM.
In addition to the above spatial semantic matching, we enhance the channel interaction via channel-wise correlation~\cite{2016CoAtt,2022COSNet}, generating $\boldsymbol{f}_{\rm dsmm}$.
Meanwhile, for the low-level features, we obtain edge information through the pooling-subtraction operation~\cite{2019AFNet}, and correct edge errors in a self-alignment way.
Then, we adopt the corrected edge features to perform feature enhancement in Edge-based Enhancement Units (EEUs) of ESAM, and also perform channel-wise correlation, generating $\boldsymbol{f}_{\rm esam}$.
The decoder of our SeaNet is comprised of three lightweight blocks denoted as D$^\textrm{1-2}$, D$^\textrm{3-4}$, and D$^\textrm{5}$.
Using $\boldsymbol{f}^{5}_{\rm e}$, $\boldsymbol{f}_{\rm dsmm}$, and $\boldsymbol{f}_{\rm esam}$, we highlight salient objects in a progressive manner for better resolution recovery.
We also introduce the saliency inference head (SalHead) after each decoder block for deep supervision~\cite{2017DSS} and final saliency map generation.

\subsection{Dynamic Semantic Matching Module}
\label{sec:DSMM}
High-level features contain affluent semantic information, which is beneficial for salient object localization.
Here, we propose DSMM to effectively activate salient regions using high-level features with limited parameters and computational cost.
Inspired by the dynamic convolution~\cite{DyConv2020}, we generate convolution kernels with existing features rather than parameter initialization to reduce the amount of parameters.
Furthermore, we extend the dynamic convolution with the depthwise separable convolution (DSconv)~\cite{MobileNet1}, and propose the dynamic depthwise convolution (DDconv) to simultaneously reduce the computational cost.
DDconv plays an important role in the spatial semantic matching of our DSMM.
However, considering only spatial interactions is not sufficient, we therefore further introduce channel interactions into DSMM to enhance channel dependencies through the channel-wise correlation~\cite{2016CoAtt,2022COSNet}.

We show the detailed structure of DSMM in the middle part of Fig.~\ref{fig:Framework}.
Our DSMM can be divided into two parts, \ie the spatial semantic matching and the channel-wise correlation.
In the following, we present DSMM based on these two parts.
Since the SKC unit generates semantic kernels for DSMM, we first describe this unit in detail.

\textit{1) SKC Unit}.
As shown in the right part of Fig.~\ref{fig:Framework}, the SKC unit compresses the semantic information of $\boldsymbol{f}^{5}_{\rm e}$ directly but effectively.
Since the SKC unit generates semantic kernels for $\boldsymbol{f}^{3}_{\rm e}$ and $\boldsymbol{f}^{4}_{\rm e}$ that are the inputs of DSMM, we first compress the channel number of $\boldsymbol{f}^{5}_{\rm e}$ by two parallel DSconv layers to fit that of $\boldsymbol{f}^{3}_{\rm e}$ and $\boldsymbol{f}^{4}_{\rm e}$, and then compress the resolution to a suitable size through two parallel adaptive average pooling layers, generating two semantic kernels $\boldsymbol{k}_{3}\!\in\!\mathbb{R}^{32\!\times\!5\!\times\!5}$ and $\boldsymbol{k}_{4}\!\in\!\mathbb{R}^{96\!\times\!5\!\times\!5}$.
We formulate the SKC unit as follows:
\begin{equation}
   \begin{aligned}
    \boldsymbol{k}_{t} =  {\rm AP}(  {\rm DSconv}(\boldsymbol{f}^{5}_{\rm e})),~~~t=3,4,
    \label{eq:1}
    \end{aligned}
\end{equation}
where ${\rm DSconv}(\cdot)$ is the $3\!\times\!3$ DSconv layer, and ${\rm AP}(\cdot)$ is the adaptive average pooling layer.

\textit{2) Spatial Semantic Matching}.
We adopt $\boldsymbol{k}_{3}$ and $\boldsymbol{k}_{4}$ containing global information as the kernel of dynamic convolution layers~\cite{DyConv2020} to reduce the amount of parameters.
However, the computational cost of traditional dynamic convolution layer is the same as that of regular convolution layer.
Inspired by DSconv~\cite{MobileNet1}, we update dynamic convolution to DDconv, which performs dynamic convolution in a depthwise manner, \ie the group of dynamic convolution is set to the channel number of input features.
Moreover, we introduce the dilation mechanism~\cite{Dila2016} into DDconv, and use multiple dilated receptive fields to sufficiently perceive salient objects for accurate localization, which is effective for overcoming the scenes of multiple objects and objects with variable sizes in ORSIs.

As shown in DSMM of Fig.~\ref{fig:Framework}, we respectively employ three dilated DDconv layers with dilation rates $\{1, 2, 3\}$ on $\boldsymbol{f}^{3}_{\rm e}$ and $\boldsymbol{f}^{4}_{\rm e}$.
In fact, DDconv is a parameter-free semantic matching process, which breaks the constraints of the trained parameters and improves the flexibility of the module and even the network.
Different from DSconv that executes depthwise convolution and pointwise convolution sequentially, we integrate the output features of three dilated DDconvs through element-wise summation, and then perform pointwise convolution to fuse the multi-perception features.
This not only further reduces the amount of parameters, but also facilitates the interaction of features generated with different receptive fields.
We formulate the spatial semantic matching as follows:
\begin{equation}
   \begin{aligned}
     \boldsymbol{f}^{t}_{\rm sm} =  {\rm Pconv}( \sum_{i=1}^3 {\rm DDconv}(\boldsymbol{f}^{t}_{\rm e}; \boldsymbol{k}_{t}, {r}_{i}) ),~~~t=3,4,
    \label{eq:2}
    \end{aligned}
\end{equation}
where $\boldsymbol{f}^{t}_{\rm sm}\!\in\!\mathbb{R}^{c_t\!\times\!h_t\!\times\!w_t}$ denotes the output features of semantic matching, ${\rm DDconv}(\cdot; \boldsymbol{k}_{t}, {r}_{i}) $ is the DDconv layer with dynamic kernel $\boldsymbol{k}_{t}$ and dilation rate ${r}_{i}\!\in\!\{1,2,3\}$, $\sum$ is the element-wise summation of multiple features, and ${\rm Pconv}(\cdot)$ is the pointwise convolution layer.

\begin{figure}
\centering
\footnotesize
  \begin{overpic}[width=0.85\columnwidth]{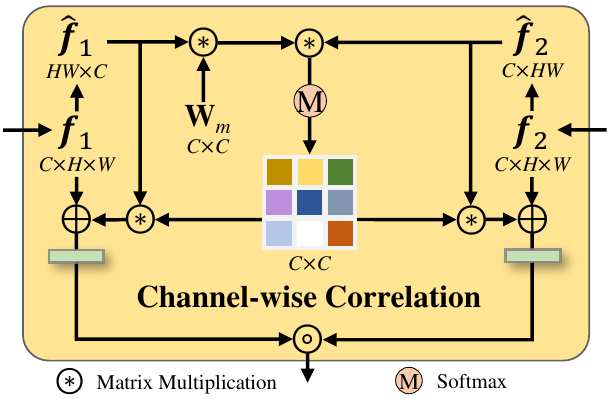}
  \end{overpic}
\caption{
Illustration of the Channel-wise Correlation.
}
\label{CCorr_structure}
\end{figure}

\textit{3) Channel-wise Correlation}.
In addition to the spatial interactions, we perform the channel-wise correlation extended from spatial co-attention~\cite{2016CoAtt,2022COSNet} in DSMM.
We first align $\boldsymbol{f}^{3}_{\rm sm}$ and $\boldsymbol{f}^{4}_{\rm sm}$ in channel and resolution via a DSconv layer and an upsampling operation to obtain the input features of channel-wise correlation, \ie $\{\boldsymbol{\hat{f}}^{3}_{\rm sm},\boldsymbol{\hat{f}}^{4}_{\rm sm}\}\!\in\!\mathbb{R}^{c_4\!\times\!h_3\!\times\!w_3}$.
Then, we define the channel-wise correlation, denoted by $\rm CCorr(\cdot)$, as follows:
\begin{equation}
   \begin{aligned}
     \boldsymbol{f}_{\rm dsmm} =  {\rm CCorr}( \boldsymbol{\hat{f}}^{3}_{\rm sm}, \boldsymbol{\hat{f}}^{4}_{\rm sm} ),
    \label{eq:3}
    \end{aligned}
\end{equation}
where $\boldsymbol{f}_{\rm dsmm} \in \mathbb{R}^{(2\!\times\!c_4)\!\times\!h_3\!\times\!w_3}$ denotes the output features of DSMM.

We depict the structure of channel-wise correlation in Fig.~\ref{CCorr_structure}, where we simplify the input features to $\{\boldsymbol{f}_{1},\boldsymbol{f}_{2}\}\!\in\!\mathbb{R}^{C\times\!H\times\!W}$ for brevity.
First, we reshape the size of $\boldsymbol{f}_{1}$ and $\boldsymbol{f}_{2}$, and obtain the flattened $\boldsymbol{\hat{f}}_{1}\!\in\!\mathbb{R}^{HW\times\!C}$ and $\boldsymbol{\hat{f}}_{2}\!\in\!\mathbb{R}^{C\times\!HW}$.
Then, we multiply $\boldsymbol{\hat{f}}_{1}$ by a trainable matrix $\mathrm{\mathbf{W}}_{m} \in \mathbb{R}^{C\times\!C}$ using matrix multiplication to adaptively learn feature transformations.
The channel-wise affinity matrix $\mathrm{\mathbf{A}}\!\in\!\mathbb{R}^{C\times\!C}$ of $\boldsymbol{\hat{f}}_{1}$ and $\boldsymbol{\hat{f}}_{2}$ can be calculated through matrix multiplication as follows:
\begin{equation}
   \begin{aligned}
     \mathrm{\mathbf{A}} =  \boldsymbol{\hat{f}}_{2} \circledast ( \boldsymbol{\hat{f}}_{1} \circledast \mathrm{\mathbf{W}}_{m} ),
    \label{eq:4}
    \end{aligned}
\end{equation}
where $\circledast$ is the matrix multiplication. In this way, we model feature dependencies along the channel.

Then, we adopt the row-wise and column-wise softmax functions to normalize the affinity matrix respectively, and transfer the established channel dependencies to $\boldsymbol{\hat{f}}_{1}$ and $\boldsymbol{\hat{f}}_{2}$.
Besides, we introduce the short connection and a $3\times3$ DSconv layer to integrate the original spatially enhanced features (\ie $\boldsymbol{f}_{1}$ and $\boldsymbol{f}_{2}$) and the above channel-enhanced features.
We formulate the above process as follows:
\begin{equation}
   \begin{aligned}
    \boldsymbol{f}^{1}_{\rm sc} =  {\rm DSconv} \big(  ( {\rm M}_r(\mathrm{\mathbf{A}}) \circledast \boldsymbol{\hat{f}}_{1}^{\top}  ) \oplus \boldsymbol{f}_{1} \big),
    \label{eq:5}
    \end{aligned}
\end{equation}
\begin{equation}
   \begin{aligned}
    \boldsymbol{f}^{2}_{\rm sc} =  {\rm DSconv} \big( (  ({\rm M}_c(\mathrm{\mathbf{A}}))^{\top}  \circledast \boldsymbol{\hat{f}}_{2} ) \oplus \boldsymbol{f}_{2} \big),
    \label{eq:6}
    \end{aligned}
\end{equation}
where $\{ \boldsymbol{f}^{1}_{\rm sc}, \boldsymbol{f}^{2}_{\rm sc} \} \in \mathbb{R}^{C\!\times\!H\!\times\!W}$ are features of spatial and channel enhancement, ${\rm M}_r(\cdot)$ and ${\rm M}_c(\cdot)$ are the row-wise and column-wise softmax functions, respectively, ${\top}$ is the matrix transpose operation, and $\oplus$ is the element-wise summation.
Notably, we omit feature size transformation for brevity.
Finally, we concatenate $\boldsymbol{f}^{1}_{\rm sc}$ and $\boldsymbol{f}^{2}_{\rm sc} $ to produce the output features of channel-wise correlation $\boldsymbol{f}_{\rm ccorr} \in \mathbb{R}^{2C\!\times\!H\!\times\!W}$, \ie the output features of DSMM $\boldsymbol{f}_{\rm dsmm}$.

In summary, our DSMM is implemented in a lightweight manner with limited parameters and computational cost.
And we fully consider spatial interactions and channel interactions in DSMM, providing comprehensive and accurate localization of salient objects, which is conducive to conquering the challenging scenes of ORSIs.

\subsection{Edge Self-Alignment Module}
\label{sec:ESAM} 
Low-level features contain rich texture and object detail information, which is conducive to delineating the fine structure of salient objects.
We propose ESAM to effectively and efficiently explore edge information for detail enhancement to preserve the complex shapes of salient objects in ORSIs.
Different from some edge-based ORSI-SOD methods~\cite{2021DAFNet,2022MCCNet,2022MJRBM,2022EMFINet}, our ESAM is lightweight, and extracts edge information without using edge supervision, which is more convenient.
Like DSMM, ESAM also fully considers the spatial interaction and channel interaction of features.
As illustrated in the left part of Fig.~\ref{fig:Framework}, ESAM consists of two Edge-based Enhancement Units (EEUs) and one channel-wise correlation.
We describe them in turn.

\textit{1) Edge-based Enhancement Unit with Self-Alignment}.
The input features of ESAM are $\boldsymbol{f}^{1}_{\rm e}$ and $\boldsymbol{f}^{2}_{\rm e}$.
We align them via DSconv layer and upsampling operation, and obtain $\{\boldsymbol{\hat{f}}^{1}_{\rm e},\boldsymbol{\hat{f}}^{2}_{\rm e}\}\!\in\!\mathbb{R}^{c_2\!\times\!h_1\!\times\!w_1}$, which are the input features of EEUs.
We adopt the pooling-subtraction operation~\cite{2019AFNet} to extract two groups of edge features $\{\boldsymbol{\hat{f}}^{1}_{\rm edge},\boldsymbol{\hat{f}}^{2}_{\rm edge}\}\!\in\!\mathbb{R}^{c_2\!\times\!h_1\!\times\!w_1}$ from $\boldsymbol{\hat{f}}^{1}_{\rm e}$ and $\boldsymbol{\hat{f}}^{2}_{\rm e}$ respectively as follows:
\begin{equation}
   \begin{aligned}
    \boldsymbol{f}^{t}_{\rm edge} =  \boldsymbol{\hat{f}}^{t}_{\rm e}  \ominus {\rm AP}( \boldsymbol{\hat{f}}^{t}_{\rm e}),~~~t=1,2,
    \label{eq:7}
    \end{aligned}
\end{equation}
where  $\ominus$ is the element-wise subtraction.
$\boldsymbol{f}^{t}_{\rm edge}$ is then used to delineate edge regions in $\boldsymbol{\hat{f}}^{t}_{\rm e}$.
However, since there is no edge supervision, the obtained edge information is inevitably vulnerable to errors.
Attacking this problem, we propose a novel self-alignment mechanism based on the mean squared error loss (\ie ${L}_{2}$  or ${L}_{mse}$ loss), and apply it to edge features, as shown in ESAM of Fig.~\ref{fig:Framework}.
In this way, we can adaptively correct edge errors between $\boldsymbol{\hat{f}}^{1}_{\rm edge}$ and $\boldsymbol{\hat{f}}^{2}_{\rm edge}$ during the training phase, and obtain accurate and consistent edge information.

Based on the corrected edge information, we perform the edge enhancement on $\boldsymbol{\hat{f}}^{t}_{\rm e}$ and obtain the output features of EEU, denoted as $\boldsymbol{f}^{t}_{\rm eeu}\!\in\!\mathbb{R}^{c_2\!\times\!h_1\!\times\!w_1}$, as follows:
\begin{equation}
   \begin{aligned}
    \boldsymbol{f}^{t}_{\rm eeu} =  {\rm conv}_{1\times1}^{s}(\boldsymbol{f}^{t}_{\rm edge}) \otimes \boldsymbol{\hat{f}}^{t}_{\rm e} \oplus \boldsymbol{\hat{f}}^{t}_{\rm e},~~~t=1,2,
    \label{eq:8}
    \end{aligned}
\end{equation}
where ${\rm conv}_{1\times1}^{s}(\cdot)$ is the $1\times1$ convolution layer with sigmoid activation function, and $\otimes$ is the element-wise multiplication.

\textit{2) Channel-wise Correlation}.
EEUs focus on feature enhancement at the spatial level, and we enhance the channel interaction of their outputs $\boldsymbol{f}^{1}_{\rm eeu}$ and $\boldsymbol{f}^{2}_{\rm eeu}$ using channel-wise correlation as follows:
\begin{equation}
   \begin{aligned}
     \boldsymbol{f}_{\rm esam} =  {\rm CCorr}( \boldsymbol{f}^{1}_{\rm eeu}, \boldsymbol{f}^{2}_{\rm eeu} ),
    \label{eq:9}
    \end{aligned}
\end{equation}
where $\boldsymbol{f}_{\rm esam}\!\in\!\mathbb{R}^{(2\!\times\!c_2)\!\times\!h_1\!\times\!w_1}$ is the output features of ESAM.

In this way, ESAM saves a large amount of parameters and computational cost, while providing powerful support for accurately highlighting the complex geometry and topology of salient objects in ORSIs.

\begin{table}[!t]
\centering
\caption{Detailed structure and parameters of three decoder blocks.
  $(3\times3, 320, 192)$ denotes that the kernel size of DSconv is $3\times3$, the input channel number is 320, and the output channel number is 192.
  }
\label{tab:Decoder_Details}
\renewcommand{\arraystretch}{1.45}
\renewcommand{\tabcolsep}{1.4mm}
\begin{tabular}{c|c|c|c}
\bottomrule
  \hline
  Aspects & D$^\textrm{5}$ &  D$^\textrm{3-4}$ & D$^\textrm{1-2}$ \\
  \hline
  \hline
    Input size    & $[320\times9\times9]$    & $[384\times36\times36]$ & $[96\times144\times144]$   \\
  \hline
    DSconv       & $(3\times3, 320, 320)$   & $(3\times3, 384, 192)$    & $(3\times3, 96, 48)$   \\
    DSconv       & $(3\times3, 320, 320)$   & $(3\times3, 192, 192)$     & $(3\times3, 48, 48)$   \\
    Upsampling & $4\times$                        & $4\times$                         & $2\times$   \\
    DSconv       & $(3\times3, 320, 192)$    & $(3\times3, 192, 48)$       & $(3\times3, 48, 48)$   \\
  \hline
    Output size  & $[192\times36\times36]$ & $[48\times144\times144]$ & $[48\times288\times288]$   \\
\toprule
\end{tabular}
\end{table}
\begin{table*}[t!]
  \centering
  \small
  \renewcommand{\arraystretch}{1.45}
  \renewcommand{\tabcolsep}{0.8mm}
  \caption{
    Quantitative and computational complexity comparisons with state-of-the-art NSI-SOD methods, ORSI-SOD methods, and lightweight methods on EORSSD and ORSSD datasets.
   $\uparrow$ indicates that the higher the better, while $\downarrow$ is the opposite.
   The top three results are marked in \textcolor{red}{\textbf{red}}, \textcolor{blue}{\textbf{blue}} and \textcolor{green}{\textbf{green}}, respectively.
    }
\label{table:QuantitativeResults}
  
\resizebox{1\textwidth}{!}{
\begin{tabular}{r|c|c|c|c|c|cccccccc|cccccccc}
\midrule[1pt]    
 \multirow{2}{*}{\normalsize{Methods}}
 & \multirow{2}{*}{\normalsize{Type}}
  & Input
& Speed
& \#Param
& FLOPs
 & \multicolumn{8}{c|}{EORSSD~\cite{2021DAFNet}} 
 & \multicolumn{8}{c}{ORSSD~\cite{2019LVNet}}  \\
  \cline{7-14} \cline{15-22} 
       &  &  size & (fps)$\uparrow$ & (M)$\downarrow$ & (G)$\downarrow$ & $S_{\alpha}\uparrow$ & $F_{\beta}^{\rm{max}}\uparrow$ & $F_{\beta}^{\rm{mean}}\uparrow$ & $F_{\beta}^{\rm{adp}}\uparrow$ & $E_{\xi}^{\rm{max}}\uparrow$ & $E_{\xi}^{\rm{mean}}\uparrow$ & $E_{\xi}^{\rm{adp}}\uparrow$ & $ \mathcal{M}\downarrow$
   	          & $S_{\alpha}\uparrow$ & $F_{\beta}^{\rm{max}}\uparrow$ & $F_{\beta}^{\rm{mean}}\uparrow$ & $F_{\beta}^{\rm{adp}}\uparrow$ & $E_{\xi}^{\rm{max}}\uparrow$ & $E_{\xi}^{\rm{mean}}\uparrow$ & $E_{\xi}^{\rm{adp}}\uparrow$ & $ \mathcal{M}\downarrow$\\
	     
\midrule[1pt]
DSS$_{17}$~\cite{2017DSS}         	& CN & 400$\times$300 & 8 & 62.23 & 114.6 & .7868 & .6849 & .5801 & .4597 & .9186 & .7631 & .6933 & .0186 
									 & .8262 & .7467 & .6962 & .6206 & .8860 & .8362 & .8085 & .0363 \\
RADF$_{18}$~\cite{2018RADF}    	& CN & 400$\times$400 & 7 & 62.54 & 214.2 & .8179 & .7446 & .6582 & .4933 & .9140 & .8567 & .7162 & .0168 
									 & .8259 & .7619 & .6856 & .5730 & .9130 & .8298 & .7678 & .0382 \\
R3Net$_{18}$~\cite{2018R3Net}   	& CN & 300$\times$300 & 2 & 56.16 &  47.5  & .8184 & .7498 & .6302 & .4165 & .9483 & .8294 & .6462 & .0171
									 & .8141 & .7456 & .7383 & .7379 & .8913 & .8681 & .8887 &  .0399\\
PoolNet$_{19}$~\cite{2019PoolNet}  & CN & 400$\times$300 & 25 & 53.63 & 123.4 & .8207 & .7545 & .6406 & .4611 & .9292 & .8193 & .6836 & .0210
									    & .8403 & .7706 & .6999 & .6166 & .9343 & .8650 & .8124 & .0358 \\ 
EGNet$_{19}$~\cite{2019EGNet}  	& CN & $\sim$380$\times$320 & 9 & 108.07 & 291.9 & .8601 & .7880 & .6967 & .5379 & .9570 & .8775 & .7566 & .0110  
									 & .8721 & .8332 & .7500 & .6452 & .9731 & .9013 & .8226 & .0216 \\ 
GCPA$_{20}$~\cite{2020GCPA}  	& CN & 320$\times$320  & 23 & 67.06  & 54.3 & .8869 & .8347 & .7905 & .6723 & .9524 & .9167 & .8647 & .0102  
									 & .9026 & .8687 & .8433 & .7861 & .9509 & .9341 & .9205 & .0168 \\ 
MINet$_{20}$~\cite{2020MINet}  	& CN & 320$\times$320 & 12 & 47.56 & 146.3 & .9040 & .8344 & .8174 & .7705 & .9442 & .9346 & .9243 & .0093
									   & .9040 & .8761 & .8574 & .8251 & .9545 & .9454 & .9423 & .0144 \\ 
ITSD$_{20}$~\cite{2020ITSD}  		& CN & 288$\times$288 & 16 & 17.08 & 54.5 & .9050 & .8523 & .8221 & .7421 & .9556 & .9407 & .9103 & .0106
									   & .9050 & .8735 & .8502 & .8068 & .9601 & .9482 & .9335 & .0165 \\ 
GateNet$_{20}$~\cite{2020GateNet} & CN & 384$\times$384 & 25 & 100.02 & 108.3 & .9114 & .8566 & .8228 & .7109 & .9610 & .9385 & .8909 & .0095
									    & .9186 & .8871 & .8679 & .8229 & .9664 & .9538 & .9428 & .0137 \\ 
SUCA$_{21}$~\cite{2021SUCA}  	& CN & 256$\times$256 & 24 & 117.71 & 56.4 & .8988 & .8229 & .7949 & .7260 & .9520 & .9277 & .9082 & .0097
									   & .8989 & .8484 & .8237 & .7748 & .9584 & .9400 & .9194 & .0145 \\
PA-KRN$_{21}$~\cite{2021PAKRN}  & CN & 600$\times$600 & 16 & 141.06 & 617.7 & .9192 & .8639 & .8358 & .7993 & .9616 & .9536 & .9416 & .0104
									   & .9239 & .8890 & .8727 & .8548 & .9680 & .9620 & .9579 & .0139 \\									   						
\hline
LVNet$_{19}$~\cite{2019LVNet}  	  & CR & 128$\times$128 & 1.4 & -   & - & .8630 & .7794 & .7328 & .6284 & .9254 & .8801 & .8445 & .0146 
									      & .8815 & .8263 & .7995 & .7506 & .9456 & .9259 & .9195 & .0207\\
DAFNet$_{21}$~\cite{2021DAFNet}    & CR & 128$\times$128 & 26 & 29.35 & 68.5 & .9166 & .8614 & .7845 & .6427 &  \textcolor{red}{\textbf{.9861}} & .9291 & .8446 & \textcolor{red}{\textbf{.0060}} 
									      & .9191 & .8928 & .8511 & .7876 & \textcolor{green}{\textbf{.9771}} & .9539 & .9360 & .0113 \\ 
SARNet$_{21}$~\cite{2021SARNet} & CR & 336$\times$336 & 47 & 25.91 & 129.7 & .9240 & .8719 & \textcolor{green}{\textbf{.8541}} & \textcolor{blue}{\textbf{.8304}} & .9620 & .9555 & .9536 & .0099
									   & .9134 & .8850 & .8619 & .8512 & .9557 & .9477 & .9464 & .0187  \\							     
MJRBM$_{22}$~\cite{2022MJRBM} & CR & 352$\times$352 & 32 & 43.54 & 95.7 & .9197 & .8656 & .8239 & .7066 & .9646 & .9350 & .8897 & .0099
									   & .9204 & .8842 & .8566 & .8022 & .9623 & .9415 & .9328 & .0163  \\
EMFINet$_{22}$~\cite{2022EMFINet} & CR & 256$\times$256 & 25 & 107.26  & 480.9 & \textcolor{blue}{\textbf{.9290}} & \textcolor{green}{\textbf{.8720}} & .8486 & .7984 & \textcolor{green}{\textbf{.9711}} & .9604 & .9501 & .0084
									     & \textcolor{green}{\textbf{.9366}} & \textcolor{green}{\textbf{.9002}} & \textcolor{green}{\textbf{.8856}} & .8617 & .9737 & .9671 & .9663 & .0109  \\
MCCNet$_{22}$~\cite{2022MCCNet} 	  & CR & 256$\times$256 & 95 &  67.65  & 112.8 & \textcolor{red}{\textbf{.9327}} & \textcolor{red}{\textbf{.8904}} & \textcolor{blue}{\textbf{.8604}} & \textcolor{green}{\textbf{.8137}} & \textcolor{blue}{\textbf{.9755}} & \textcolor{red}{\textbf{.9685}} & \textcolor{green}{\textbf{.9538}} & \textcolor{blue}{\textbf{.0066}}
				       & \textcolor{red}{\textbf{.9437}} &  \textcolor{red}{\textbf{.9155}} & \textcolor{red}{\textbf{.9054}} & \textcolor{red}{\textbf{.8957}} & \textcolor{red}{\textbf{.9800}} & \textcolor{red}{\textbf{.9758}} & \textcolor{red}{\textbf{.9735}} & \textcolor{red}{\textbf{.0087}} \\	
\hline
CSNet$_{20}$~\cite{2020CSNet} 	& LN & 224$\times$224 & 38 & 0.14 & 0.7 & .8364 & .8341 & .7656 & .6319 & .9535 & .8929 & .8339 & .0169
									   & .8910 & .8790 & .8285 & .7615 & .9628 & .9171 & .9068 & .0186 \\
SAMNet$_{21}$~\cite{2021SAMNet} 	& LN & 336$\times$336 & 44 & 1.33 & 0.5 & .8622 & .7813 & .7214 & .6114 & .9421 & .8700 & .8284 & .0132
									   & .8761 & .8137 & .7531 & .6843 & .9478 & .8818 & .8656 & .0217 \\	
HVPNet$_{21}$~\cite{2021HVPNet} 	& LN & 336$\times$336 & 26 & 1.23 & 1.1 & .8734 & .8036 & .7377 & .6202 & .9482 & .8721 & .8270 & .0110
									   & .8610 & .7938 & .7396 & .6726 & .9320 & .8717 & .8471 & .0225 \\								   
\hline
CorrNet$_{22}$~\cite{2022CorrNet} 		& LR & 256$\times$256 &  100 &  4.09  & 21.1 & \textcolor{green}{\textbf{.9289}} & \textcolor{blue}{\textbf{.8778}} & \textcolor{red}{\textbf{.8620}} & \textcolor{red}{\textbf{.8311}} & .9696 & \textcolor{green}{\textbf{.9646}} & \textcolor{blue}{\textbf{.9593}} & .0083
									   &  \textcolor{blue}{\textbf{.9380}} &  \textcolor{blue}{\textbf{.9129}} & \textcolor{blue}{\textbf{.9002}} & \textcolor{blue}{\textbf{.8875}} & \textcolor{blue}{\textbf{.9790}} & \textcolor{blue}{\textbf{.9746}} & \textcolor{blue}{\textbf{.9721}} & \textcolor{blue}{\textbf{.0098}}  \\	

\hline
\hline
\textbf{SeaNet (Ours)}				 & LR & 288$\times$288 &  96 &  2.76  & 1.7 & .9208 & .8649 & .8519 & \textcolor{blue}{\textbf{.8304}} & .9710 & \textcolor{blue}{\textbf{.9651}} & \textcolor{red}{\textbf{.9602}} & \textcolor{green}{\textbf{.0073}} 
									   & .9260 & .8942 & .8772 &  \textcolor{green}{\textbf{.8625}} & .9767 &  \textcolor{green}{\textbf{.9722}} &  \textcolor{green}{\textbf{.9670}} &  \textcolor{green}{\textbf{.0105}}  \\		
									   							   								   
\toprule[1pt]
\multicolumn{21}{l}{\small{CN: CNN-based NSI-SOD method, CR: CNN-based ORSI-SOD method, LN: lightweight NSI-SOD method, LR: lightweight ORSI-SOD method.}} \\
\end{tabular}
}
\end{table*}

\subsection{Decoder and Loss Function}
\label{sec:Decoder_and_LossFunction}
\textit{1) Decoder}.
Based on the locations and details of salient objects provided by the above two modules, we design a lightweight decoder to produce saliency maps.
As shown at the bottom of Fig.~\ref{fig:Framework}, our lightweight decoder consists of three blocks, \ie D$^\textrm{1-2}$, D$^\textrm{3-4}$, and D$^\textrm{5}$.
Each decoder block in turn contains two DSconv layers, an upsampling operation, and another DSconv layer.
Their detailed parameters are reported in Tab.~\ref{tab:Decoder_Details}.
In particular, we arrange SalHeads after these three decoder blocks to generate three saliency maps of different resolutions, \ie $\mathbf{S}^{\textrm{3}} \in [0,1]^{1\!\times\!36\!\times\!36}$, $\mathbf{S}^{\textrm{2}} \in [0,1]^{1\!\times\!144\!\times\!144}$, and $\mathbf{S}^{\textrm{1}} \in [0,1]^{1\!\times\!288\!\times\!288}$, where the first two are used for deep supervision and the last one is the final output of our SeaNet.
SalHead is comprised of a dropout layer~\cite{2014dropout} and a $1\times1$ convolution layer.

\textit{2) Loss Function}.
We impose the binary cross-entropy (BCE) loss and intersection-over-union (IoU) loss to jointly train our SeaNet.
Therefore, our total loss consists of two parts, \ie the saliency loss and the edge alignment loss.
Moreover, we introduce a loss weight to treat these two losses differently for better training.
The total loss function ${L}_{\rm total}$ can be formulated as follows:
\begin{equation}
   \begin{aligned}
    {L}_{\rm total}  =  \sum_{i=1}^3 ( {L}_{bce}^i  + {L}_{iou}^i  ) + \lambda \cdot {L}_{mse}({\rm P}(\boldsymbol{f}^{1}_{\rm edge}),{\rm P}(\boldsymbol{f}^{2}_{\rm edge})),
    \label{eq:SalLoss}
    \end{aligned}
\end{equation}
where ${L}_{bce}^i$ and ${L}_{iou}^i$ are BCE loss and IoU loss, respectively, which supervise $\mathbf{S}^{i}$ by the ground truth (GT); $\lambda$ is the loss weight and set to 0.5; ${L}_{mse}(\cdot)$ is the mean squared error loss; and ${\rm P}(\cdot)$ is the parametric rectified linear unit~\cite{InitialWei}. 

\section{Experiments}
\label{sec:exp}

\subsection{Experimental Setup}
\label{sec:ExpProtocol}
\textit{1) Datasets.}
We conduct experiments on the ORSSD~\cite{2019LVNet} and EORSSD~\cite{2021DAFNet} datasets.
The ORSSD dataset has 800 ORSIs and corresponding pixel-level annotations.
It is divided into two parts, \ie the training set (600 images) and the test set (200 images).
The EORSSD dataset adds 1200 ORSIs to the ORSSD dataset, resulting in 2000 ORSIs.
This dataset is also divided into two parts, \ie 1400 images in the training set and 600 images in the test set.

\textit{2) Evaluation Metrics.}
We adopt quantitative evaluation metrics and computational complexity metrics to evaluate our method and all compared methods from two aspects.

Quantitative evaluation metrics include
S-measure ($S_{\alpha}$, $\alpha$ = 0.5)~\cite{Fan2017Smeasure},
F-measure ($F_{\beta}$, $\beta^{2}$ = 0.3)~\cite{Fmeasure},
E-measure ($E_{\xi}$)~\cite{Fan2018Emeasure}, and
mean absolute error ($\mathcal{M}$).
The first three are the higher the better, and the last one is the opposite.
We report the maximum, mean, and adaptive F-measure and E-measure.

Computational complexity metrics include the inference speed measured in frames per second (fps), the parameter amount (\#Param) measured in million (M), and the number of floating point operations (FLOPs) measured in giga (G).
The first one is the higher the better, and the last two are the opposite.
The inference speed is reported with a batch size of 1 and no I/O time.

\textit{3) Training Protocol.}
We conduct experiments based on the PyTorch~\cite{PyTorch} on a computer with an NVIDIA Titan X GPU (12GB memory).
We list the training details and parameters as follows:
the input size is 288$\times$288,
the data augmentation strategy includes flipping and rotation,
the network optimizer is Adam~\cite{Adam},
the batch size is 8,
the base learning rate is $1e^{-4}$,
the learning rate decays to 1/10 every 30 epochs,
and the training epoch is 50.
Besides, we initialize MobileNet-V2 with the pre-trained parameters, and initialize the newly added layers of two modules and decoder with the ``Kaiming" method~\cite{InitialWei}.
Notably, for each dataset, we train on its own training set and test on its own test set, as in~\cite{2021DAFNet,2022EMFINet,2022CorrNet}.

\subsection{Performance Analysis}
We compare our SeaNet with 17 state-of-the-art conventional SOD methods,
including 11 conventional NSI-SOD methods (\ie DSS~\cite{2017DSS}, RADF~\cite{2018RADF}, R3Net~\cite{2018R3Net}, PoolNet~\cite{2019PoolNet}, EGNet~\cite{2019EGNet}, GCPA~\cite{2020GCPA}, MINet~\cite{2020MINet}, ITSD~\cite{2020ITSD}, GateNet~\cite{2020GateNet}, SUCA~\cite{2021SUCA}, and PA-KRN~\cite{2021PAKRN}),
and 6 conventional ORSI-SOD methods (\ie LVNet~\cite{2019LVNet}, DAFNet~\cite{2021DAFNet}, SARNet~\cite{2021SARNet}, MJRBM~\cite{2022MJRBM}, EMFINet~\cite{2022EMFINet}, and MCCNet~\cite{2022MCCNet}).
In addition, we also compare our lightweight SeaNet with 4 state-of-the-art lightweight SOD methods, including 3 lightweight NSI-SOD methods (\ie CSNet~\cite{2020CSNet}, SAMNet~\cite{2021SAMNet}, and HVPNet~\cite{2021HVPNet}), and the lightweight ORSI-SOD method CorrNet~\cite{2022CorrNet}.
The above NSI-SOD methods are retrained on the same ORSI-SOD datasets as our SeaNet with default parameter settings to generate saliency maps.
We obtain saliency maps for other methods from authors or public source codes.

\textit{1) Comparison with Conventional SOD Methods.}
The top of Tab.~\ref{table:QuantitativeResults} shows the quantitative evaluation and computational complexity evaluation results of our SeaNet and conventional SOD methods for NSIs and ORSIs.
Compared with conventional NSI-SOD methods, our SeaNet achieves the best performance in terms of both accuracy and computational complexity.
For example, compared with the best performing NSI-SOD solution PA-KRN~\cite{2021PAKRN}, SeaNet has significant advantages on $\mathcal{M}$, \eg 0.0073 v.s. 0.0104 on the EORSSD dataset and 0.0139 v.s. 0.0105 on the ORSSD dataset, and has 6$\times$ faster inference speed, 51.1$\times$ fewer parameters, and 363.3$\times$ fewer FLOPs than it.
This shows the advantages of specialized methods, even our lightweight specialized method can outperform conventional NSI-SOD solutions.

Compared with conventional ORSI-SOD methods, our SeaNet shows comparable accuracy but with significantly lower computational complexity.
For example, compared with EMFINet~\cite{2022EMFINet}, SeaNet achieves similar accuracy, \eg $E_{\xi}^{\rm max}$: 0.9710 v.s. 0.9711 on the EORSSD dataset and 0.9767 v.s. 0.9737 on the ORSSD dataset, while SeaNet is 3.6$\times$ faster in inference speed, and 24.5$\times$ and 282.8$\times$ fewer in parameters and FLOPs, respectively.
Compared with the best performing MCCNet~\cite{2022MCCNet}, although the accuracy of SeaNet is not dominant, the lower computational complexity of SeaNet is still outstanding.

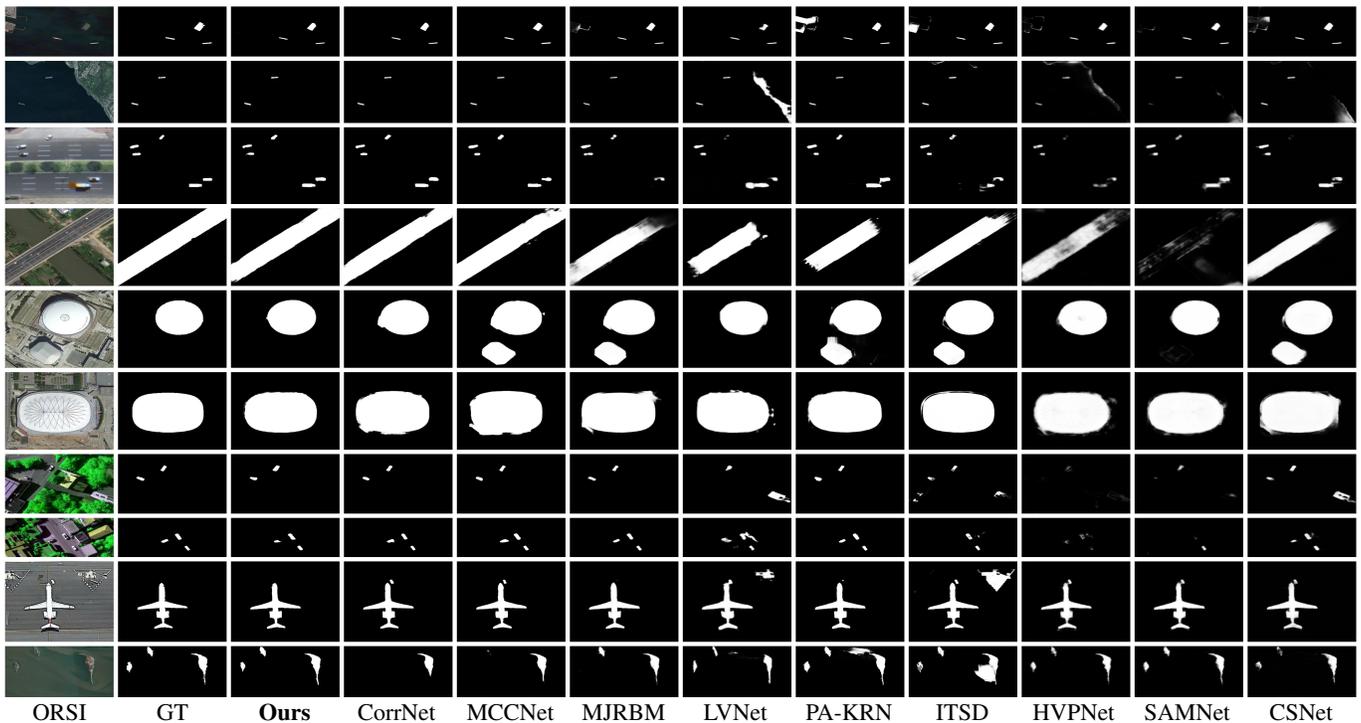
\begin{figure*}[t!]
    \centering
    \small
	\begin{overpic}[width=1\textwidth]{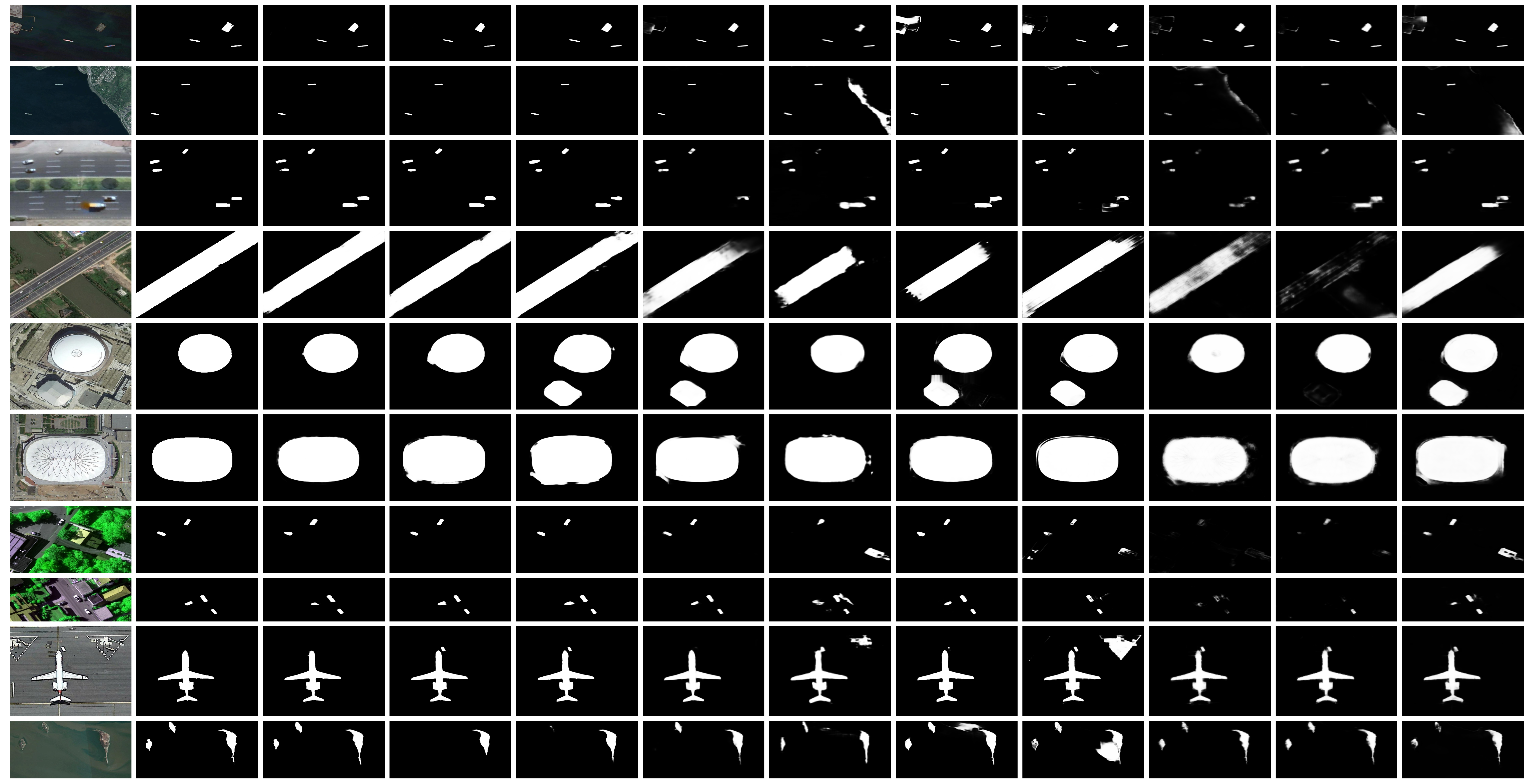}

    \put(2.,-1.35){ ORSI }
    \put(11.11,-1.35){ GT}
    \put(18.60,-1.35){ \textbf{Ours}}
    \put(25.84,-1.35){ CorrNet  }
    \put(33.83,-1.35){ MCCNet }
    \put(42.28,-1.35){ MJRBM }
    \put(51.19,-1.35){ LVNet }
    \put(58.69,-1.35){ PA-KRN }
    \put(68.23,-1.35){ ITSD }
    \put(75.33,-1.35){ HVPNet }
    \put(83.5,-1.35){ SAMNet }
    \put(92.68,-1.35){ CSNet }
    
    \end{overpic}
	\caption{Qualitative comparisons with four types of methods, including nine representative state-of-the-art methods.
    }
    \label{fig:VisualExample}
\end{figure*}

\textit{2) Comparison with Lightweight SOD Methods.}
At the bottom of Tab.~\ref{table:QuantitativeResults}, we report the comparison results of our SeaNet with four lightweight SOD methods for NSIs and ORSIs.
Compared with lightweight NSI-SOD methods, SeaNet has no advantages in parameters and FLOPs, but has obvious advantages in inference speed.
This means that our lightweight SeaNet has room for improvement.
On the other hand, the accuracy advantage of SeaNet is obvious, \eg SeaNet outperforms them by 3.50\%$\sim$8.44\% in $S_{\alpha}$, 4.87\%$\sim$13.76\% in $F_{\beta}^{\rm mean}$, 5.51\%$\sim$10.05\% in $E_{\xi}^{\rm mean}$, and 0.0037$\sim$0.0120 in $\mathcal{M}$ on two datasets.

The original intention of our SeaNet is to reduce the amount of parameters and FLOPs of existing lightweight ORSI-SOD method CorrNet~\cite{2022CorrNet}, especially the latter one.
The results in Tab.~\ref{table:QuantitativeResults} show that SeaNet achieves this goal without significantly reducing accuracy.
Specifically, in terms of computational complexity, SeaNet has 1.3$\times$ fewer parameters and 12.4$\times$ fewer FLOPs than CorrNet, and has a comparable inference speed.
In terms of accuracy, SeaNet shows a slightly lower $E_{\xi}^{\rm adp}$ (0.9670 v.s. 0.9721) than CorrNet on the ORSSD dataset, while achieves a slightly higher $\mathcal{M}$ (0.0073 v.s. 0.0083) on the EORSSD dataset.

Overall, SeaNet achieves one first place, two second places, and five third places in quantitative evaluation metrics with much more desirable computational complexity, striking a balance between effectiveness and efficiency.
This means that SeaNet is a promising method that can be applied in practical applications.

\textit{3) Qualitative Comparison.}
Here, we show the qualitative comparison of our SeaNet and four types of methods, including nine representative state-of-the-art methods, on four unique and challenging ORSI scenes in Fig.~\ref{fig:VisualExample}.
The first scene contains multiple objects or even multiple tiny objects, as shown in the first three cases of Fig.~\ref{fig:VisualExample}.
We can observe that the saliency maps of our SeaNet are similar to those of specialized ORSI-SOD methods, such as CorrNet and MCCNet, which accurately highlight all salient objects, and are much better than those of conventional and lightweight NSI-SOD methods.
This is attributed to the multi-scale semantic matching of DSMM.
The second scene contains a big object, such as the 4th to 6th cases of Fig.~\ref{fig:VisualExample}.
Most methods can locate the big object, but cannot highlight them completely, while our SeaNet can highlight the entire big object with complete edges.
This is attributed to the edge-based detail enhancement of ESAM.
The third scene contains chaotic background, such as the 7th and 8th cases of Fig.~\ref{fig:VisualExample}.
Our SeaNet can successfully find salient objects in complex backgrounds, while the three lightweight methods either miss objects (HVPNet and SAMNet) or introduce background regions (CSNet).
The last scene contains objects with complex geometric shapes, such as the last two cases of Fig.~\ref{fig:VisualExample}.
Thanks to the cooperation between DSMM and ESAM, our SeaNet has obvious advantages compared to the nine methods, that is, it can not only overcome the interference of the small object, but also precisely locate three islands with fine details.
Overall, our SeaNet can catch up with specialized ORSI-SOD methods and outperform NSI-SOD methods.

\begin{table}[!t]
\centering
\caption{Ablation results of evaluating the contribution of two lightweight modules.
  The best one in each column is \textbf{bold}.
  }
\label{Ab_two_module}
\renewcommand{\arraystretch}{1.4}
\renewcommand{\tabcolsep}{0.95mm}
\begin{tabular}{c||cc|cccc}
\bottomrule
 \multirow{2}{*}{Models}  & \#Param & FLOPs & \multicolumn{4}{c}{EORSSD~\cite{2021DAFNet}}  \\
 \cline{4-7}
   & (M)$\downarrow$ & (G)$\downarrow$ & $S_{\alpha}\uparrow$ & $F_{\beta}^{\rm{mean}}\uparrow$ 
    & $E_{\xi}^{\rm{mean}}\uparrow$ & $ \mathcal{M}\downarrow$ \\
\hline
\hline
SeaNet (\textbf{Ours}) & 2.76 & 1.66 & \textbf{0.9208} & \textbf{0.8519} & \textbf{0.9651}  & \textbf{0.0073} \\

\hline
\textit{w/o DSMM} & 2.75 \textbf{\tiny{-0.01}} & 1.56 \textbf{\tiny{-0.10}}  & 0.9158 & 0.8440 & 0.9580  & 0.0093  \\
\textit{w/o ESAM} & 2.70 \textbf{\tiny{-0.06}}  & 1.59 \textbf{\tiny{-0.07}} & 0.9180 & 0.8404 & 0.9588  & 0.0084  \\ 

\toprule
\end{tabular}
\end{table}

\subsection{Ablation Studies}
\label{Ablation Studies}
To evaluate the effectiveness of each component of our SeaNet, we conduct exhaustive ablation studies on the EORSSD dataset.
Specifically, we analyze
1) the contribution of two lightweight modules,
2) the effectiveness of each component of ESAM, and
3) the effectiveness of each component of DSMM.
The parameter settings and datasets for each variant are the same as in Sec.~\ref{sec:ExpProtocol}.

\textit{1) Contribution of two lightweight modules}.
To analyze the contribution of two lightweight modules, we provide two variants:
1) removing DSMM and SKC (\ie \textit{w/o DSMM}) and 2) removing ESAM (\ie \textit{w/o ESAM}). 
We report the quantitative results and computational complexity in Tab.~\ref{Ab_two_module}.

We observe that our two modules are lightweight, \ie the combination of DSMM and SKC have 0.01M parameters and 0.10G FLOPs, and ESAM has 0.06M parameters and 0.07G FLOPs.
Since DSMM can determine the location of salient objects, \textit{w/o DSMM} reduces the accuracy of object localization, resulting in a drastic drop in pixel-level evaluation metric, \ie $\mathcal{M}$: 0.0073$\rightarrow$0.0093.
\textit{w/o EASM} only affects the details of salient objects, so the performance degradation is generally less severe than that of \textit{w/o DSMM}, \eg $\mathcal{M}$: 0.0073$\rightarrow$0.0084 and $E_{\xi}^{\rm{mean}}$: 0.9651$\rightarrow$0.9588.
The cooperation of these two lightweight modules and MobileNet-V2 enables our SeaNet to achieve good performance without many parameters and computational cost.

\begin{table}[!t]
\centering
\caption{Ablation results of embedding the two lightweight modules into HVPNet and SAMNet.
  }
\label{Embed_into_HVPSAM}
\renewcommand{\arraystretch}{1.4}
\renewcommand{\tabcolsep}{0.8mm}
\begin{tabular}{c||cc|cccc}
\bottomrule
 \multirow{2}{*}{Models}  & \#Param & FLOPs & \multicolumn{4}{c}{EORSSD~\cite{2021DAFNet}}  \\
 \cline{4-7}
   & (M)$\downarrow$ & (G)$\downarrow$ & $S_{\alpha}\uparrow$ & $F_{\beta}^{\rm{mean}}\uparrow$ 
    & $E_{\xi}^{\rm{mean}}\uparrow$ & $ \mathcal{M}\downarrow$ \\
\hline
\hline
SeaNet (\textbf{Ours}) & 2.76 & 1.7 & 0.9208 & 0.8519 & 0.9651 & 0.0073 \\

\hline
SAMNet                & 1.33  & 0.5 & 0.8622 & 0.7214 & 0.8700  & 0.0132  \\
SeaNet-SAM & 1.51  & 1.4 & 0.9173 & 0.8507 & 0.9631  & 0.0074  \\ 

\hline
HVPNet                & 1.23  & 1.1 & 0.8734 & 0.7377 & 0.8721  & 0.0110 \\
SeaNet-HVP & 1.48  & 1.8 & 0.9202 & 0.8486 & 0.9646  & 0.0069 \\

\toprule
\end{tabular}
\end{table}

%
In particular, to further illustrate the effectiveness, flexibility, and robustness of these two lightweight modules, we embed these two lightweight modules and our decoder into the feature extraction backbones proposed by SAMNet and HVPNet, forming two variants, named SeaNet-SAM and SeaNet-HVP, respectively.
Notably, our decoder has 0.47M parameters and 0.95G FLOPs.
As shown in Tab.~\ref{Embed_into_HVPSAM}, SeaNet-SAM is more lightweight than our original SeaNet.
SeaNet-HVP has significantly fewer parameters and slightly higher FLOPs than our original SeaNet.
Both variants have comparable performance as our original SeaNet, which shows that these two lightweight modules can be adapted to different backbones and are robust.
In addition, SeaNet-SAM/SeaNet-HVP has similar number of parameters as SAMNet/HVPNet, while has higher FLOPs which are mainly from the decoder.
The performance of SeaNet-SAM/SeaNet-HVP is significantly better than that of SAMNet/HVPNet, such as leading by more than 10\% on $F_{\beta}^{\rm{mean}}$.
Notably, the number of parameters and FLOPs of SeaNet-SAM (1.51M and 1.4G) and HVPNet (1.23M and 1.1G) are comparable, while SeaNet-SAM outperforms HVPNet by a large margin.
This situation proves that when the number of parameters and FLOPs of our SeaNet variants are equivalent or comparable to those of other lightweight methods, our SeaNet variants still significantly outperform them.

\textit{2) Effectiveness of each component of DSMM.}
To analyze the effectiveness of each component of DSMM, we design three variants of DSMM in Tab.~\ref{Ab_component} and embed them into the network:
1) removing the spatial semantic matching (\ie \textit{w/o SM}),
2) changing dilated DDconvs to regular DDconvs (\ie \textit{w/o dilation}), and
3) removing the channel-wise correlation of DSMM (\ie \textit{w/o CCorr1}).

Based on the quantitative performance at the top of Tab.~\ref{Ab_two_module}, we observe that each component of DSMM is necessary.
As the key part of DSMM, \textit{w/o SM} truncates the object localization capability of DSMM, achieving the worst performance among these three variants, \eg $E_{\xi}^{\rm{mean}}$: 0.9651$\rightarrow$0.9595.
\textit{w/o dilation} impairs the ability of DSMM to perceive salient objects of different sizes, resulting a slight performance drop, \eg a drop of 0.29\% on $F_{\beta}^{\rm{mean}}$.
\textit{w/o CCorr1} only focuses on the feature interactions at the spatial level, while ignoring that at the channel level, resulting in incomplete feature interaction and performance degradation.
Therefore, the above analysis proves that the design of our DSMM is reasonable and effective.

\begin{table}[!t]
\centering
\caption{Ablation results of evaluating the effectiveness of each component of DSMM and ESAM.
  The best one in each column is \textbf{bold}.
  }
\label{Ab_component}
\renewcommand{\arraystretch}{1.4}
\renewcommand{\tabcolsep}{2.2mm}
\begin{tabular}{l|c||cccc}
\bottomrule
  & \multirow{2}{*}{Models}  & \multicolumn{4}{c}{EORSSD~\cite{2021DAFNet}}  \\
 \cline{3-6}
  &  & $S_{\alpha}\uparrow$ & $F_{\beta}^{\rm{mean}}\uparrow$ 
    & $E_{\xi}^{\rm{mean}}\uparrow$ & $ \mathcal{M}\downarrow$ \\
\hline
\hline
& SeaNet (\textbf{Ours}) & \textbf{0.9208} & \textbf{0.8519} & \textbf{0.9651}  & \textbf{0.0073} \\

\hline
  \multirow{3}{*}{\begin{sideways}DSMM\end{sideways}}
& \textit{w/o SM}       & 0.9170 & 0.8459 & 0.9595  & 0.0078   \\ 
& \textit{w/o dilation}       & 0.9194 & 0.8490 & 0.9626  & 0.0076   \\
& \textit{w/o CCorr1}       & 0.9187 & 0.8471 & 0.9619  & 0.0080   \\

\hline
  \multirow{3}{*}{\begin{sideways}EASM\end{sideways}}
& \textit{w/o EEU}              & 0.9183 & 0.8441 & 0.9597  & 0.0078  \\
& \textit{w/o alignment}      & 0.9189 & 0.8476 & 0.9617  & 0.0078  \\
& \textit{w/o CCorr2}         & 0.9197 & 0.8494 & 0.9627  & 0.0076  \\ 

\toprule
\end{tabular}
\end{table}

\textit{3) Effectiveness of each component of EASM.}
To analyze the effectiveness of each component of EASM, we design three variants of EASM in Tab.~\ref{Ab_component} and embed them into the network:
1) removing two EEUs (\ie \textit{w/o EEU}),
2) removing the edge alignment (\ie removing the edge alignment loss in Eq.~\ref{eq:SalLoss}, named \textit{w/o alignment}), and
3) removing the channel-wise correlation of EASM (\ie \textit{w/o CCorr2}).

According to the quantitative performance at the bottom of Tab.~\ref{Ab_two_module}, we observe that each component of EASM contributes to the final performance.
The two EEUs are responsible for enhancing the edge regions on the features, so \textit{w/o EEU} does not achieve satisfactory performance, \eg $F_{\beta}^{\rm{mean}}$: 0.8519$\rightarrow$0.8441.
The proposed edge alignment is an interesting mechanism that can improve the accuracy of edge information without increasing parameters and FLOPs.
\textit{w/o alignment} causes performance degradation on all four metrics.
Like \textit{w/o CCorr1}, \textit{w/o CCorr2} also gives up the channel-level feature interactions in EASM, which hurts performance.
Combining \textit{w/o CCorr1} and \textit{w/o CCorr2}, we can conclude that channel-level feature interactions are important to our SeaNet and cannot be discarded.
The above analysis shows that these components of ESAM are indispensable.

\section{Conclusion}
\label{sec:con}
In this paper, we aim to treat low-level and high-level features discriminatively, thereby proposing an efficient solution, named SeaNet, for lightweight ORSI-SOD.
For high-level features, the lightweight DSMM is proposed to explore object locations through spatial semantic matching, and takes into account the channel-level feature interactions.
Spatial semantic matching utilizes dilated DDconvs to perceive multiple salient objects and objects with variable sizes, resulting in good adaptation to complex scenes of ORSIs.
For low-level features, the lightweight ESAM is proposed to enhance the details of salient objects based on edge information which is corrected in an innovative self-alignment manner.
With the close cooperation of the two modules, SeaNet infers salient objects accurately in the decoder at a fast speed.
Performance analysis and ablation studies demonstrate the effectiveness and efficiency of our SeaNet compared with state-of-the-art methods.



\ifCLASSOPTIONcaptionsoff
  \newpage
\fi

\bibliographystyle{IEEEtran}
\bibliography{SeaNetRef}

%



%

\end{document}